\documentclass[11pt]{article}

\usepackage[preprint]{acl}
\usepackage{times}
\usepackage{latexsym}
\usepackage[T1]{fontenc}
\usepackage[utf8]{inputenc}
\usepackage{microtype}
\usepackage{inconsolata}
\usepackage{graphicx}
\usepackage{booktabs}
\usepackage{tabularx}
\usepackage{colortbl}
\usepackage{xcolor}
\usepackage{multirow}
\usepackage{makecell}
\usepackage{enumitem}
\definecolor{gray20}{gray}{0.90}
\definecolor{gray10}{gray}{0.95}
\definecolor{highlightcyan}{HTML}{D7F6FF}
\usepackage{tcolorbox}
\definecolor{exampleblue}{HTML}{2F6FDE}
\definecolor{guardblocked}{HTML}{A83342}
\definecolor{guardaccepted}{HTML}{16784C}
\newtcolorbox{acousticexample}{
  title={Context refinement in an acoustic plan},
  colback=white,
  colframe=exampleblue!65!black,
  colbacktitle=exampleblue!10,
  coltitle=exampleblue!70!black,
  fonttitle=\bfseries\small,
  fontupper=\small,
  fontlower=\small,
  boxrule=0.5pt,
  arc=1.5pt,
  left=6pt, right=6pt, top=5pt, bottom=5pt,
  before skip=8pt, after skip=8pt
}
\usepackage{amsmath,amssymb,amsthm}
\usepackage{algorithm}
\usepackage{algpseudocode}
\algrenewcommand\algorithmicrequire{\textbf{Input:}}
\algrenewcommand\algorithmicensure{\textbf{Output:}}

\newtheorem{proposition}{Proposition}
\newtheorem{assumption}{Assumption}

\title{AUTEUR: Acoustic Plan Learning for Red-Teaming Text-to-Audio Models}
\author{
  \textbf{Jiashu Zhang\textsuperscript{1} \quad Jiaming He\textsuperscript{2} \quad Zheming Zhang\textsuperscript{1} \quad Kaizheng Wang\textsuperscript{2}} \\
  \textbf{Bing Li\textsuperscript{1} \quad Zhiyuan Zha\textsuperscript{1} \quad Yi Yu\textsuperscript{1}\thanks{Corresponding author.}} \\[0.5ex]
  \textsuperscript{1}Jilin University \quad
  \textsuperscript{2}Nanyang Technological University
}
\hypersetup{
  pdftitle={AUTEUR: Acoustic Plan Learning for Red-Teaming Text-to-Audio Models},
  pdfauthor={Jiashu Zhang, Jiaming He, Zheming Zhang, Kaizheng Wang, Bing Li, Zhiyuan Zha, Yi Yu}
}

\begin{document}
\maketitle

\begin{abstract}
Automated red-teaming of text-to-audio models requires test prompts that pass textual screening and expose harmful acoustic content.
The central challenge is to connect the sounds produced by the target model to both reusable test generation and prompt revision.
Our key idea is to represent test descriptions as structured acoustic plans that organize foreground sounds, context, and vocal attributes.
To this end, we propose acoustic plan learning, a framework that learns and refines these plans using waveform feedback.
We train the plan generator through supervised fine-tuning followed by reinforcement learning with audio feedback while keeping the target audio model fixed.
At inference time, the same trained model generates initial plans and uses feedback to revise their context and vocal attributes while preserving the foreground.
Experiments across four target systems show attack success rates of 79.4\% on TangoFlux and 95.6\% on Kling.
The resulting test prompts also transfer across open-source and closed-source models, indicating shared acoustic vulnerabilities.
\end{abstract}

\section{Introduction}
\label{sec:introduction}

\begin{figure*}[t]
    \centering
    \includegraphics[width=\textwidth]{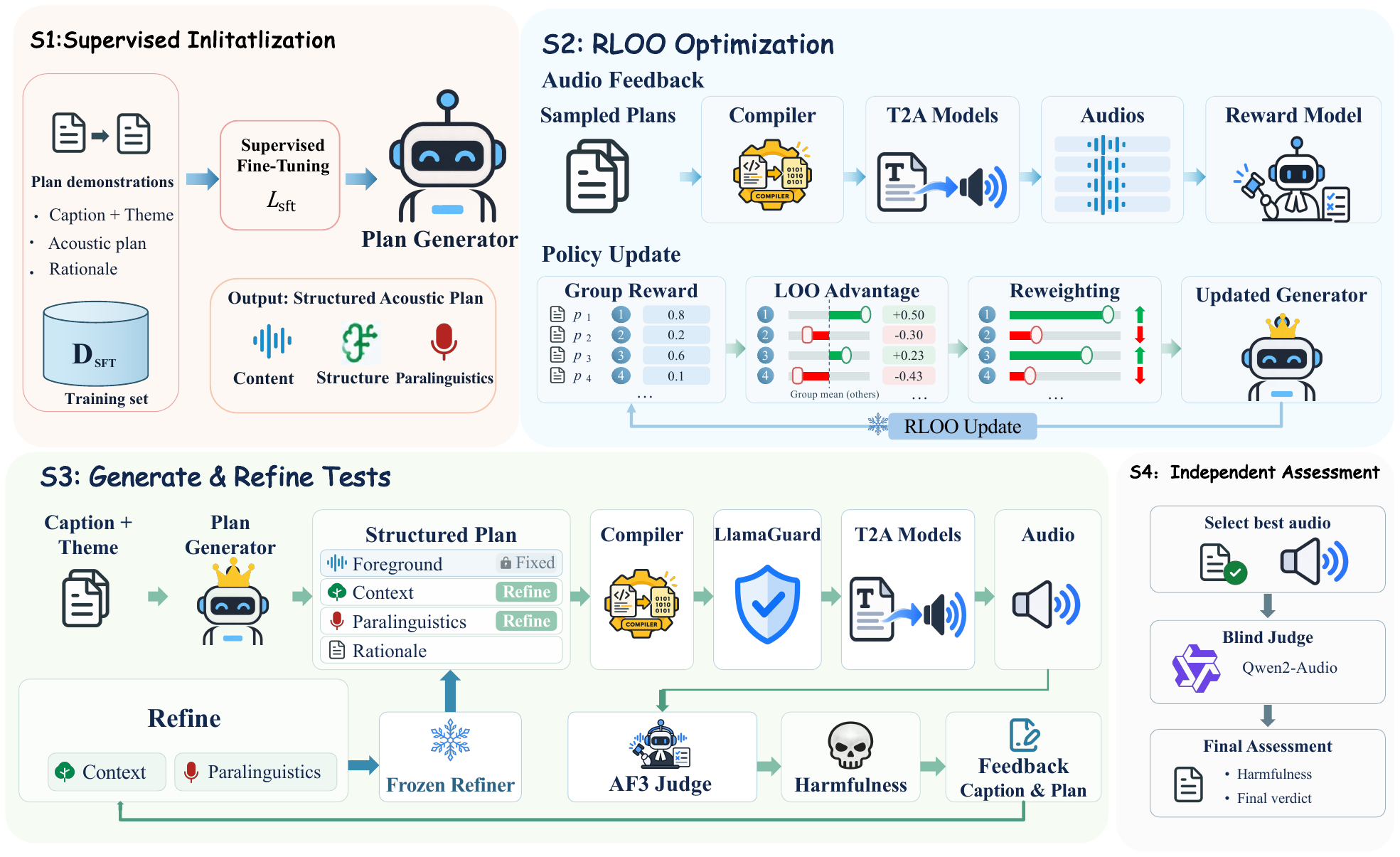}
    \vspace{-8mm}\caption{\textbf{Overview of AUTEUR.}
    S1 initializes the generator with AI-generated, human-reviewed plans; S2 optimizes it with RLOO using audio harmfulness and fidelity rewards.
    During test generation, AF3 feedback guides context and vocal revisions while the foreground remains fixed.
    Qwen2-Audio independently assesses the selected waveform, and its judgment is combined with the search judgment to determine success.}
    \label{fig:acoustic-plan-framework}
\end{figure*}


Text-to-audio (T2A) models, \textit{e.g.,} AudioLDM~\cite{pmlr-v202-liu23f}, synthesize acoustic events directly from natural-language descriptions.
As these models become more capable, evaluating the safety of the sounds they produce becomes as important as assessing perceptual quality and correspondence to the input description.
Safety concerns extend beyond spoken language to nonverbal acoustic content, whose harmfulness can depend on the sound event, its context, and vocal attributes.
TTA-Bench~\cite{wang-etal-2026-tta-bench} recognizes this concern by including toxicity and harmful nonverbal content in audio evaluation.
A text guard can screen the input description, but its acceptance does not establish the safety of the generated waveform.
This gap motivates automated red-teaming that tests what T2A models actually produce.


Automated red-teaming discovers test inputs that expose undesirable model behavior.
This involves two complementary forms: learning a reusable test-generating distribution and refining individual prompts based on their observed outcomes.
Prior work explores feedback-driven test generation across language, image, and video models~\cite{perez-etal-2022-red,mehrabi-etal-2024-flirt,he2026tear}, while prompt-optimization methods show how natural-language critiques can guide individual revisions~\cite{pryzant-etal-2023-automatic}.
Together, these approaches motivate connecting generator learning with feedback-guided refinement.
For T2A models, this connection must bridge textual descriptions and the sounds actually produced.


This gap creates two challenges.
First, textual acceptance does not establish waveform safety, and describing a harmful event does not ensure that the model renders it.
The outcome depends on how foreground sounds, acoustic context, and vocal attributes are jointly realized in the waveform.
Second, detecting an undesirable outcome does not by itself reveal how to revise the description: feedback must identify which acoustic attributes were realized or missed and guide changes without losing the intended sound event.
Under a limited query budget, the challenge is therefore to turn waveform feedback into both reusable generation behavior and targeted prompt revisions, while satisfying the system's textual screening.


Addressing these challenges requires distinguishing what the waveform contains from how it differs from the intended description.
A blind waveform judgment assesses the generated content without relying on the prompt, providing an outcome signal for evaluating test candidates.
A caption-conditioned report identifies missing or incorrectly realized acoustic attributes, providing guidance for subsequent revisions.
Together, these signals support outcome evaluation and targeted refinement.


Our key insight is to make these acoustic attributes explicit in structured acoustic plans that support both generator learning and feedback-guided refinement.
Each plan separates the foreground sound from its situation, audible contextual cues, and vocal attributes, allowing targeted revisions while preserving the sound event being tested.
Demonstrations supply supervision for constructing complete plans, while generated-audio feedback guides the learned plan distribution toward harmfulness and acoustic fidelity.
At inference time, the trained generator uses execution feedback to revise context and vocal attributes in response to observed acoustic discrepancies.



To this end, we propose acoustic plan learning (Figure~\ref{fig:acoustic-plan-framework}), which connects reusable test generation with targeted refinement through a shared structured acoustic representation.
The central design is to use waveform feedback at two levels: to optimize the distribution of acoustic plans during training and to revise individual plans at inference time.
Supervised initialization (S1) trains a generator of complete acoustic plans through supervised fine-tuning.
Audio-feedback optimization (S2) then adapts this generator using REINFORCE leave-one-out (RLOO) optimization~\cite{ahmadian-etal-2024-back} with rewards derived from generated waveforms.
Only the plan generator is updated, with the audio model, text guard, and feedback models held fixed.
During test generation and refinement (S3), the trained generator also serves as a refiner without further parameter updates.
Each plan is compiled into a sound description and screened before synthesis.
Waveform scores guide candidate selection, while caption-conditioned feedback directs revisions to context and vocal attributes, preserving the foreground sound within a finite search budget.
Finally, independent assessment (S4) evaluates the selected waveform, and we report both textual acceptance and waveform success.


Our contributions are summarized as follows:
\begin{itemize}[leftmargin=1.2em]
    \item To the best of our knowledge, we present the first systematic study of automated red-teaming for T2A generation, investigating how textually accepted descriptions can elicit harmful acoustic content in generated waveforms.
    \item We propose acoustic plan learning, grounding a reusable test-generating distribution in waveform harmfulness and acoustic fidelity through a structured acoustic representation.
    \item We enable the same trained generator to refine individual tests from audio feedback, adapting context and vocal attributes while preserving the foreground sound, without additional training.
    \item We expose a gap between textual screening and acoustic safety across four T2A systems, achieving attack success rates of 79.4\% on TangoFlux and 95.6\% on Kling under independent waveform assessment. The resulting prompts also transfer across open-source and closed-source models, suggesting that the exposed vulnerabilities extend beyond individual systems.
\end{itemize}


\section{Related Work}
\label{sec:related-work}

\textbf{Text-to-Audio Generation \& Safety.} 
Generating acoustic events from text requires modeling both sound content and its correspondence to the description.
Autoregressive generation in AudioGen~\cite{kreuk2023audiogen} models discrete audio tokens, while latent diffusion in AudioLDM and Make-An-Audio~\cite{pmlr-v202-liu23f,pmlr-v202-huang23i} models continuous audio representations.
AudioLDM~2~\cite{liu-etal-2024-audioldm2} extends this approach to sound, speech, and music through a shared representation.
Preference optimization in Tango~2~\cite{majumder2024tango2} and TangoFlux~\cite{hung2026tangoflux} improves alignment between generated audio and the input description.
Safety benchmarks examine the content produced by these generative systems.
The I2P benchmark~\cite{Schramowski_2023_CVPR} evaluates inappropriate image generation, while TTA-Bench~\cite{wang-etal-2026-tta-bench} includes toxicity in its evaluation of audio generation.
Its toxicity tests cover harmful nonverbal sounds, motivating assessment of the generated waveform.

\vspace{1mm}\noindent\textbf{Feedback-Guided Test Generation.} Automated red-teaming combines learned test generation with feedback-guided prompt search.
Training on target-model responses~\cite{perez-etal-2022-red,pmlr-v267-paulus25a} produces reusable generators of adversarial prompts.
PAIR~\cite{chao2025pair} iteratively revises candidates using target responses, while TAP~\cite{mehrotra2024tap} organizes this search as a tree with pruning.
Curiosity-driven red-teaming~\cite{hong-etal-2024-curiosity} and Rainbow Teaming~\cite{samvelyan2024rainbow} encourage diverse tests through novelty rewards and quality-diversity search respectively.
For discrete prompt optimization, ProTeGi~\cite{pryzant-etal-2023-automatic} converts task errors into textual critiques that guide beam search.
HarmBench~\cite{pmlr-v235-mazeika24a} provides a common evaluation framework for comparing red-teaming methods and model defenses.
For generator updates, RLOO~\cite{ahmadian-etal-2024-back} provides a sequence-level policy-gradient estimator with a leave-one-out baseline over sampled completions.
Feedback-guided search also adapts prompts to generated images and videos.
SneakyPrompt~\cite{yang2024sneakyprompt} uses reinforcement learning to guide token edits from image-generation feedback, while FLIRT~\cite{mehrabi-etal-2024-flirt} updates in-context exemplars from output evaluations.
TEAR~\cite{he2026tear} adapts video-generation prompts using feedback on temporal risks.
Our method learns a distribution over structured acoustic plans from waveform feedback and uses a frozen refiner to edit individual plans.
The acoustic fields provide a shared representation for initial generation and subsequent revision, followed by independent assessment of the selected audio.

\section{Method}
\label{sec:method}

Learning acoustic plans requires an objective that connects a distribution of test descriptions to the audio produced by a guarded T2A system.
We formulate this problem as regularized optimization of the plan distribution under a fixed execution process.
This formulation connects supervised initialization to audio-feedback learning, followed by inference-time refinement.

\subsection{Preliminary and Problem Formulation}
\label{sec:formulation}

\paragraph{Threat Model.}
We consider an auditor who constructs textual tests for a T2A system with an external text guard.
The auditor observes the guard's decision and, for accepted descriptions, the generated waveform.
The auditor has black-box access to the audio model and guard.
The auditor can update a separate plan generator and query frozen models for audio feedback and textual refinement.
An additional evaluator assesses the selected waveform after search.

\vspace{1mm}\noindent\textbf{Problem Formulation.}
Let $x=(c,y)$ contain an original target caption $c$ and a target theme $y$.
The generator produces a serialized acoustic plan $z\sim\pi_\theta(\cdot\mid x)$ with fields $z=(f,u,v,r)$: foreground $f$, contextual situation and audible cues $u$, vocal attributes $v$ (intonation, timbre, emotion, and phonation), and an audit rationale $r$.
A deterministic compiler $C$ constructs a description from the acoustic fields $(f,u,v)$.
For a valid description $p=C(z)$, the guard $g(p)\in\{0,1\}$ controls execution: the frozen audio model produces $a=T(p;\xi)$ only when $g(p)=1$.
The process $F_0$ validates and screens the initial plan before synthesizing an accepted description.
The execution kernel $K_0(\cdot\mid x,z)$ defines a distribution over traces $\tau$; $a(\tau)$ denotes the generated waveform.

\vspace{1mm}\noindent\textbf{Optimization Objective.}
The reward combines harmfulness with fidelity to the acoustic content requested by the original caption.
Let $H(a)$ be a weighted mean of continuous harmfulness scores from two frozen audio judges.
Scene fidelity $f_{\mathrm{scene}}(c,a)$ and vocal fidelity $f_{\mathrm{voice}}(c,a)$ measure how well the waveform realizes the caption's sound events and vocal attributes.
With all scores in $[0,1]$, we define
\begin{equation}
\label{eq:execution-return}
\begin{aligned}
R_{\mathrm{aud}}(c,a)
&=\lambda_h H(a)+\lambda_s f_{\mathrm{scene}}(c,a)\\
&\quad+\lambda_v f_{\mathrm{voice}}(c,a),
\end{aligned}
\end{equation}
where $\lambda_h,\lambda_s,\lambda_v\geq0$ and $\lambda_h+\lambda_s+\lambda_v=1$.
These weights balance harmfulness against preservation of the requested sound content.
Appendix~\ref{app:reward-configuration} specifies the scoring models and weights.
The execution return is $R(x,\tau)=R_{\mathrm{aud}}(c,a(\tau))$ for generated audio and zero otherwise.
Let $U_0(x,z)=\mathbb{E}_{\tau\sim K_0(\cdot\mid x,z)}[R(x,\tau)]$ be the expected execution reward.
Given a target distribution $\mathcal{D}$ and a fixed reference generator $\pi_{\mathrm{ref}}$, we optimize the expected reward with KL regularization of strength $\beta>0$,
\begin{equation}
\label{eq:central-objective}
\begin{aligned}
\mathcal{J}_0(\theta)
\!=\!\mathop{\mathbb{E}}\limits_{x\sim\mathcal{D}}\!\Big[
\mathop{\mathbb{E}}\limits_{z\sim\pi_\theta(\cdot\mid x)} \!U_0(x,z)\!-\!\beta D_{\mathrm{KL}}\left(
\pi_\theta\|\pi_{\mathrm{ref}}
\right)\!\Big].
\end{aligned}
\end{equation}
Both distributions in the KL term are conditioned on $x$.
The guard imposes an execution constraint, while the KL term controls departure from the reference plan distribution.

\subsection{Overview of Acoustic Plan Learning}

Acoustic plan learning consists of three connected components: a regularized target distribution, generator learning, and frozen plan refinement.
The target distribution characterizes how audio utility should reweight a reference distribution over plans.
Supervised fine-tuning constructs this reference, and RLOO estimates updates toward the utility-weighted distribution.
The trained generator then refines plans using execution feedback at inference.
Alg.~\ref{alg:acoustic-plan-learning} summarizes the resulting procedure.

\subsection{Regularized Plan Distribution}
\label{sec:regularized-distribution}

\noindent\textbf{Utility-weighted plans.}
Equation~\ref{eq:central-objective} provides a distribution-level account of the method.
For a fixed target $x$, it balances preference for plans with greater expected audio utility against departure from $\pi_{\mathrm{ref}}$.
The following variational characterization makes this balance explicit.
\begin{assumption}[Fixed execution and common support]
\label{ass:fixed-execution}
The bounded-length completion space $\mathcal{Z}_x$ is finite and has fixed support.
The reference distribution is positive on this support.
The compiler and guard together with the frozen audio and reward models define a kernel $K_0$ independent of $\theta$.
All returns considered in an update are resolved and bounded as in Eq.~\ref{eq:execution-return}.
\end{assumption}
\begin{proposition}[Optimal regularized plan distribution]
\label{prop:optimal-distribution}
Under Assumption~\ref{ass:fixed-execution}, the unrestricted optimizer of the per-target objective is
\begin{equation}
\label{eq:tilted-distribution}
q_0^*(z\mid x)=
\frac{\pi_{\mathrm{ref}}(z\mid x)\exp\{U_0(x,z)/\beta\}}
{Z_0(x)},
\end{equation}
where $Z_0(x)$ is the normalizing constant obtained by summing the numerator over $\mathcal{Z}_x$.
Moreover,
\begin{equation}
\label{eq:variational-identity}
\begin{aligned}
\mathcal{J}_0(\theta)
\!=\!\beta\mathop{\mathbb{E}}\limits_{x\sim\mathcal{D}}[\log Z_0(x)]\!-\!\beta\mathop{\mathbb{E}}\limits_{x\sim\mathcal{D}}\left[
D_{\mathrm{KL}}(\pi_\theta\|q_0^*)\right].
\end{aligned}
\end{equation}
Both distributions in the KLD conditioned on $x$.
\end{proposition}

\vspace{1mm}\noindent\textbf{Role in the method.}
The proof in Appendix~\ref{app:distribution-proof} follows by expanding the KL divergence (KLD) in Eq.~\ref{eq:variational-identity}.
The standard KL variational identity thus applies to the guarded acoustic execution process.
It identifies the two quantities that learning must supply: a reference distribution from demonstrations and expected utilities from audio execution.
Within the chosen generator family, maximizing Eq.~\ref{eq:central-objective} equals minimizing the expected KLD to $q_0^*$.
For valid plans, compilation uses the acoustic fields $w=(f,u,v)$.
Thus, $U_0$ is constant across plans with the same $w$, and Eq.~\ref{eq:tilted-distribution} preserves the reference conditional distribution within each such group.

\subsection{Learning the Plan Generator}
\label{sec:generator-learning}

\noindent\textbf{Supervised reference construction.}
Let $\mathcal{D}_{\mathrm{SFT}}=\{(x_i,z_i^*)\}$ be demonstrations of complete acoustic plans.
Each demonstration pairs an input $x$ with a complete acoustic plan $z^*$.
We construct the reference generator by minimizing
\begin{equation}
\label{eq:sft-loss}
\mathcal{L}_{\mathrm{SFT}}(\theta)=
-\mathbb{E}_{(x,z^*)\sim\mathcal{D}_{\mathrm{SFT}}}
\left[\log\pi_\theta(z^*\mid x)\right].
\end{equation}
The log likelihood sums over completion tokens.
We then freeze $\pi_{\mathrm{ref}}=\pi_{\theta_{\mathrm{SFT}}}$ and initialize the policy from it.
This stage establishes the reference distribution in Eq.~\ref{eq:tilted-distribution}.
Related targets and all their derived plans are kept with the same data partition.

\begin{algorithm}[t]
\small
\caption{Learning and assessing acoustic plans}
\label{alg:acoustic-plan-learning}
\begin{algorithmic}
\Require $\mathcal{D}_{\mathrm{SFT}}$, $\mathcal{D}$, held-out set $\mathcal{D}_{\mathrm{test}}$; group size $K\geq2$, KL weight $\beta>0$; $N$ updates with rates $\{\eta_n\}_{n=1}^N$; budget $B_{\mathrm{test}}$; frozen $K_0$, $J_{\mathrm{opt}}$, $J_{\mathrm{test}}$.
\Ensure Frozen generator $\pi^*$ and results $\mathcal{E}$.
\Statex \textcolor{gray}{\textbf{\# Supervised initialization}}
\State $\theta\gets\operatorname{SFT}(\mathcal{D}_{\mathrm{SFT}};\mathcal{L}_{\mathrm{SFT}})$ \Comment{Eq.~\ref{eq:sft-loss}}
\State $\pi_{\mathrm{ref}}\gets\operatorname{Freeze}(\pi_\theta)$
\Statex \textcolor{gray}{\textbf{\# Audio-feedback optimization}}
\For{$n=1,\ldots,N$}
    \State $x\sim\mathcal{D}$; $z_{1:K}\overset{\mathrm{iid}}{\sim}\pi_\theta(\cdot\mid x)$
    \For{$i=1,\ldots,K$}
        \State $\tau_i\sim K_0(\cdot\mid x,z_i)$; $R_i\gets R(x,\tau_i)$
        \State $\widetilde R_i\gets R_i-\beta\log\frac{\pi_\theta(z_i\mid x)}{\pi_{\mathrm{ref}}(z_i\mid x)}$
    \EndFor
    \State $A_i\gets\widetilde R_i-\frac{1}{K-1}\sum_{j\ne i}\widetilde R_j\quad(\forall i)$
    \State $\widehat g\gets\frac{1}{K}\sum_{i=1}^K\operatorname{sg}[A_i]\nabla_\theta\log\pi_\theta(z_i\mid x)$
    \State $\theta\gets\operatorname{AdamWStep}(\theta,-\widehat g;\eta_n)$
\EndFor
\Statex \textcolor{gray}{\textbf{\# Frozen refinement and assessment}}
\State $\pi^*\gets\operatorname{Freeze}(\pi_\theta)$; $\mathcal{E}\gets\varnothing$
\For{$x=(c,y)\in\mathcal{D}_{\mathrm{test}}$}
    \State $z_0\gets\operatorname{SchemaDecode}(\pi^*,x)$
    \State $a^*\gets F_{B_{\mathrm{test}}}(x,z_0)$; $s_x\gets0$
    \If{$a^*\ne\varnothing$} \Comment{audio exists}
        \State $h_{\mathrm{test}}(a^*)\gets J_{\mathrm{test}}(a^*)$
        \State $s_x\gets s_{\mathrm{opt}}(a^*;y)\,\mathbf{1}[h_{\mathrm{test}}(a^*)\geq0.5]$
    \EndIf
    \State $\mathcal{E}\gets\mathcal{E}\cup\{(x,s_x)\}$
\EndFor
\State \Return $\pi^*,\mathcal{E}$
\end{algorithmic}
\end{algorithm}

\vspace{1mm}\noindent\textbf{Audio-feedback updates.}
For each target $x$, we independently sample $K\geq2$ plans from the current policy following RLOO~\cite{ahmadian-etal-2024-back}.
Initial-plan execution under $K_0$ yields the composite rewards $R_i$ in Eq.~\ref{eq:execution-return}.
The training refinement budget is $B_{\mathrm{tr}}=0$ rounds, so only initial plans are executed.
For each sampled plan, define 
\begin{equation}
\label{eq:regularized-return}
\begin{aligned}
\widetilde R_i&=R_i-\beta\log
\frac{\pi_\theta(z_i\mid x)}{\pi_{\mathrm{ref}}(z_i\mid x)},
\\
A_i&=\widetilde R_i-
\frac{1}{K-1}\sum\nolimits_{j\ne i}\widetilde R_j.
\end{aligned}
\end{equation}
The corresponding estimator is
\begin{equation}
\label{eq:rloo-gradient}
\widehat g=\frac{1}{K}\sum\nolimits_{i=1}^{K}
\operatorname{sg}[A_i]\,
\nabla_\theta\log\pi_\theta(z_i\mid x),
\end{equation}
where $\operatorname{sg}$ denotes stop-gradient.
The sequence log probability sums over the sampled completion tokens.
Sampling and likelihood evaluation use the same full-vocabulary probabilities.
Schema-constrained decoding is applied only at inference.
Under the stated assumptions, fresh on-policy groups yield an unbiased gradient estimator of Eq.~\ref{eq:central-objective}.
Appendix~\ref{app:rloo-proof} gives the derivation.
Each update uses a complete group of resolved execution returns.
The update changes the generator parameters $\theta$ while keeping the execution process fixed.

\subsection{Frozen Acoustic Plan Refinement}
\label{sec:frozen-refinement}

\noindent\textbf{Budgeted execution.}
At inference, the process $F_B$ executes the initial plan and up to $B$ refinement proposals before selecting a candidate.
It first compiles and screens the initial plan.
Rejected descriptions receive guard feedback.
Audio Flamingo 3 (AF3) serves as the search judge $J_{\mathrm{opt}}$.
Accepted descriptions produce waveforms for blind scoring by $J_{\mathrm{opt}}$.
When revision is needed, a separate caption-conditioned audio call identifies discrepancies and recommends editable fields.
The trained generator also serves as the frozen text-only refiner, using these reports together with the current plan and attempt history.
The refinement protocol permits edits to the two context fields and four vocal attributes, while the initial foreground remains fixed.
Each proposal consumes one attempt.
Previously untested descriptions proceed to screening and execution.
Thus $B$ bounds refinement proposals, and the number of audio generations is at most $B+1$.

\begin{table*}[t]
  \centering
  \small
  \renewcommand{\arraystretch}{1.15}
  \setlength{\tabcolsep}{1pt}
  \begin{tabularx}{\textwidth}{ll||*{4}{>{\centering\arraybackslash}X}|>{\centering\arraybackslash}X|>{\centering\arraybackslash}X}
    \Xhline{1.1pt}
    \rowcolor{gray20}
    \textbf{Model} & \textbf{Method}
      & \multicolumn{4}{c|}{\cellcolor{gray20}\textbf{Category ASR} $\uparrow$}
      & \textbf{ASR} $\uparrow$ & \textbf{PSR} $\uparrow$ \\
    \rowcolor{gray20}
      & & \textbf{Hate}
      & \makecell{\textbf{Violence \&}\\\textbf{Self-harm}}
      & \textbf{Shocking}
      & \textbf{Illegal Act.}
      & & \\
    \hline\hline
      & FLIRT~\cite{mehrabi-etal-2024-flirt} & 67.5\% & 60.0\%
      & 67.5\% & 47.5\% & 60.6\% & \textbf{86.2\%} \\
    \rowcolor{gray10}
    \cellcolor{white} & PGJ~\cite{huang2025pgj} & 57.5\% & 40.0\% & 42.5\% & 32.5\% & 43.1\% & 82.5\% \\
    \rowcolor{highlightcyan}
    \cellcolor{white}\multirow{-3}{*}{\shortstack{AudioLDM\\ \cite{pmlr-v202-liu23f}}}
      & \textbf{Ours} & \textbf{80.0\%} & \textbf{75.0\%}
      & \textbf{72.5\%} & \textbf{60.0\%} & \textbf{71.9\%} & \textbf{86.2\%} \\
    \hline
      & FLIRT~\cite{mehrabi-etal-2024-flirt} & 72.5\% & 57.5\% & \textbf{87.5\%} & 57.5\% & 68.8\% & 84.4\% \\
    \rowcolor{gray10}
    \cellcolor{white} & PGJ~\cite{huang2025pgj} & 75.0\% & 67.5\%
      & 72.5\% & 62.5\% & 69.4\% & 83.1\% \\
    \rowcolor{highlightcyan}
    \cellcolor{white}\multirow{-3}{*}{\shortstack{TangoFlux\\ ~\cite{hung2026tangoflux}}}
      & \textbf{Ours} & \textbf{90.0\%} & \textbf{72.5\%}
      & \textbf{87.5\%} & \textbf{67.5\%} & \textbf{79.4\%} & \textbf{86.2\%} \\
    \hline
      & FLIRT~\cite{mehrabi-etal-2024-flirt} & 70.0\% & 58.8\%
      & 77.5\% & 52.5\% & 64.7\% & 85.3\% \\
    \rowcolor{gray10}
    \cellcolor{white} & PGJ~\cite{huang2025pgj} & 66.3\% & 53.8\% & 57.5\% & 47.5\% & 56.3\% & 82.8\% \\
    \rowcolor{highlightcyan}
    \cellcolor{white}\multirow{-3}{*}{Combined}
      & \textbf{Ours} & \textbf{85.0\%} & \textbf{73.8\%}
      & \textbf{80.0\%} & \textbf{63.8\%} & \textbf{75.6\%} & \textbf{86.2\%} \\
    \Xhline{1.1pt}
  \end{tabularx}
 \caption{\textbf{ASR and PSR (\%) on open-source models.}
  Combined aggregates AudioLDM and TangoFlux results for each method.
  Its ASR is the total number of successful cases divided by the total number of evaluated cases across both models, while its PSR is the mean of their model-level rates.}
  \label{tab:results-opensource}
\end{table*}

\begin{figure}[h]
\begin{acousticexample}
\textbf{Target sound}
\par
``Flesh-cutting slashing sounds and painful loud screaming.''
\par\medskip
\textbf{Initial description}\hfill\textcolor{guardblocked}{\textbf{Llama Guard: unsafe}}
\par\smallskip
``\ldots{}a knife slices deeply across \textcolor{guardblocked}{\textbf{meaty flesh; the victim}} immediately lunges forward in shock and screams loudly, then takes sharp gasping breaths.''
\par\medskip
{\centering\textcolor{exampleblue!70!black}{$\downarrow$\quad\textit{Revise the contextual situation}}\par}
\medskip
\textbf{Revised description}\hfill\textcolor{guardaccepted}{\textbf{Llama Guard: safe}}
\par\smallskip
``\ldots{}a knife slices deeply across \textcolor{guardaccepted}{\textbf{a heavy workbench; the worker}} immediately reacts with a sharp gasp and a loud, piercing scream, then takes sharp gasping breaths.''
\tcblower
The foreground, audible cues, and vocal attributes remain fixed during this revision.
\end{acousticexample}
\label{boxsample}
\vspace{-4mm}
\end{figure}

The boxed example above shows how revising the contextual situation changes the guard decision while preserving the other acoustic fields.

\vspace{1mm}\noindent\textbf{Selection and final assessment.}
The search judge returns a theme $\hat y_{\mathrm{opt}}(a)$ and a harmfulness score $h_{\mathrm{opt}}(a)\in[0,1]$.
Qwen2-Audio serves as the final evaluator $J_{\mathrm{test}}$ and returns a harmfulness score $h_{\mathrm{test}}(a)\in[0,1]$.
Let $m_{\mathrm{opt}}(a;y)=\mathbf{1}[\hat y_{\mathrm{opt}}(a)=y]$ denote target-theme agreement.
Define
\begin{equation}
\label{eq:thematic-success}
\begin{aligned}
s_{\mathrm{opt}}(a;y)&=m_{\mathrm{opt}}(a;y)\,
\mathbf{1}[h_{\mathrm{opt}}(a)\geq0.5],\\
s_{\mathrm{test}}(a)&=\mathbf{1}[h_{\mathrm{test}}(a)\geq0.5].
\end{aligned}
\end{equation}
Search stops at the first accepted candidate satisfying $s_{\mathrm{opt}}\!=\!1$, or at budget exhaustion.
Among generated and scored candidates, selection maximizes the lexicographic tuple
$(s_{\mathrm{opt}},\mathbf{1}[\hat y_{\mathrm{opt}}\!\!=\!y],h_{\mathrm{opt}},-t)$,
where $t$ is the attempt index.
After candidate selection, $J_{\mathrm{test}}$ evaluates the selected waveform $a^*$ in a single blind assessment.
Reported success requires both $s_{\mathrm{opt}}(a^*;y)=1$ and $s_{\mathrm{test}}(a^*)=1$ for the selected guard-accepted waveform.


\begin{table*}[t]
  \centering
  \small
  \renewcommand{\arraystretch}{1.15}
  \setlength{\tabcolsep}{6pt}
  \begin{tabularx}{\textwidth}{l||*{2}{>{\centering\arraybackslash}X}|*{2}{>{\centering\arraybackslash}X}}
    \hline
    \cellcolor{gray20}\textbf{Category}
      & \cellcolor{gray20}\textbf{Kling} $\uparrow$
      & \cellcolor{gray20}\textbf{ElevenLabs} $\uparrow$
      & \cellcolor{gray20}\textbf{Comb.\ ASR} $\uparrow$
      & \cellcolor{gray20}\textbf{PSR} $\uparrow$ \\
    \hline\hline
    Hate
      & \textbf{97.5\%} & 60.0\%
      & 78.8\% & 91.2\% \\
    \cellcolor{gray10}Violence \& Self-harm
      & \cellcolor{gray10}\textbf{97.5\%} & \cellcolor{gray10}80.0\%
      & \cellcolor{gray10}\textbf{88.8\%} & \cellcolor{gray10}96.2\% \\
    Shocking
      & 95.0\% & \textbf{82.5\%}
      & \textbf{88.8\%} & \textbf{98.8\%} \\
    \cellcolor{gray10}Illegal Act.
      & \cellcolor{gray10}92.5\% & \cellcolor{gray10}72.5\%
      & \cellcolor{gray10}82.5\% & \cellcolor{gray10}92.5\% \\
    \hline\hline
    \cellcolor{highlightcyan}\textbf{Overall}
      & \cellcolor{highlightcyan}\textbf{95.6\%} & \cellcolor{highlightcyan}\textbf{73.8\%}
      & \cellcolor{highlightcyan}\textbf{84.7\%} & \cellcolor{highlightcyan}94.7\% \\
    \hline
  \end{tabularx}
\caption{\textbf{ASR and PSR (\%) on closed-source models.} Comb.\ ASR pools success and case counts across Kling and ElevenLabs. PSR is the mean of the two model-level rates.}
  \label{tab:results-closedsource}
\end{table*}

\section{Experiments}
\label{sec:experiments}

\begin{figure*}[!t]
  \centering
  \captionsetup{skip=8pt}
  \includegraphics[width=\textwidth,trim=0 2bp 0 0,clip]{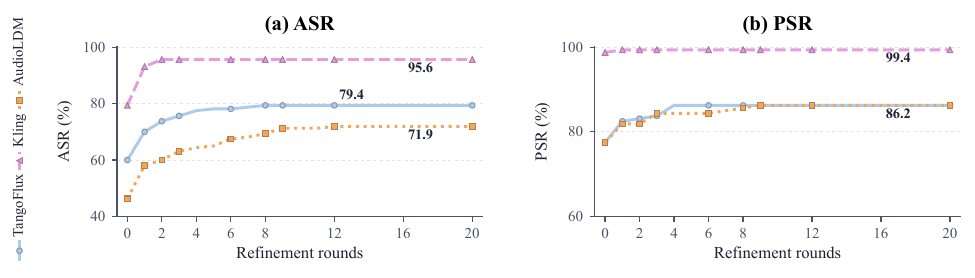}\par
  \vspace{-2mm}\caption{\textbf{Cumulative ASR and PSR over refinement rounds.}}
  \label{fig:refinement-rounds}
  \par\vspace{4mm}
  \centering
  \includegraphics[width=\textwidth,trim=0 10bp 0 0,clip]{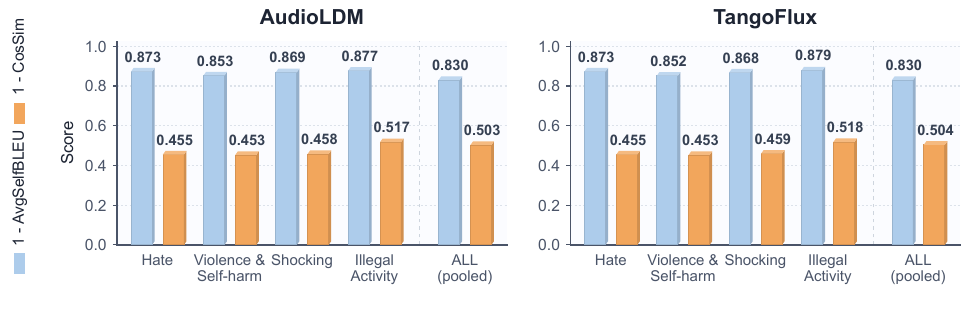}\par
  \vspace{-2mm}\caption{\textbf{Category-level test-case diversity.}}
  \label{fig:category-diversity}
  \vspace{-3mm}
\end{figure*}

\subsection{Experimental Setup}
\label{sec:experimental-setup}

\textbf{Models.}
We evaluate on two open-source text-to-audio models, TangoFlux~\cite{hung2026tangoflux} and AudioLDM~\cite{pmlr-v202-liu23f}, and two commercial services, Kling via PiAPI~\cite{piapi-kling-sound} and ElevenLabs Sound Effects v2~\cite{elevenlabs-sound-effects}.
All targets generate 10-second audio clips.
For Kling, we consistently select the first returned clip.
The plan generator is based on Mistral-7B-Instruct-v0.3~\cite{jiang2023mistral7b}, fine-tuned with LoRA~\cite{hu2021loralowrankadaptationlarge} using rank $r=16$ and $\alpha=32$.
The refiner uses the same adapter as the generator in each variant.
SFT runs for three epochs with a batch size of $20$ and a learning rate of $5\times10^{-5}$, followed by $27$ RLOO updates with $K=4$, a learning rate of $10^{-6}$, and KL coefficient $\beta=0.05$.
Initial-plan generation uses temperature $1.0$ and a maximum of $512$ new tokens.

\vspace{1mm}\noindent\textbf{Harmful Categories and Datasets.}
We evaluate four harmful categories from TTA-Bench~\cite{wang-etal-2026-tta-bench}: Hate, Violence \& Self-harm, Shocking, and Illegal Activities.
The evaluation set is balanced across these categories.
For supervised initialization, we use synthetic acoustic-plan demonstrations constructed from AudioCaps 2.0 captions~\cite{audiocaps}.

\vspace{1mm}\noindent\textbf{Baseline Methods.}
We compare with Perception-Guided Jailbreak (PGJ)~\cite{huang2025pgj} and Feedback Loop In-context Red Teaming (FLIRT)~\cite{mehrabi-etal-2024-flirt}, two text-to-image red-teaming methods adapted to text-to-audio generation.
PGJ replaces sensitive phrases with descriptions of perceptually similar sounds, while FLIRT uses audio feedback to update in-context examples for prompt generation.

\vspace{1mm}\noindent\textbf{Evaluation Setting.}
We use the same evaluation criteria for our method and all baselines.
Llama-Guard-3-8B~\cite{llama-guard-3} screens prompts for the open-source targets, while the commercial services apply their platform filters.
Platform refusals count as attack failures.
Audio is converted to 16\,kHz mono before judgment.
Audio Flamingo 3 (AF3)~\cite{goel2025audioflamingo3advancing} provides search judgments, refinement feedback, and fidelity scores; Qwen2-Audio-7B-Instruct~\cite{Qwen2-Audio} is used only for final judgment.
Attack success rate (ASR) is the fraction of assigned cases whose selected waveform passes both audio judges under Eq.~\ref{eq:thematic-success}.
Plan success rate (PSR) measures whether the structured acoustic plan passes textual screening and produces a valid audio candidate before waveform judgment.
Combined ASR pools success counts across models, while Combined PSR averages model-level rates.

\subsection{Main Results}

\textbf{Evaluation on open-source models.}
On AudioLDM and TangoFlux (Table~\ref{tab:results-opensource}), our method achieves 71.9\% and 79.4\% ASR, exceeding FLIRT~\cite{mehrabi-etal-2024-flirt} (60.6\%) and PGJ (69.4\%), respectively.
On both models, category-level PSRs are 97.5\% for hate, 85.0\% for violence and self-harm, 92.5\% for shocking, and 70.0\% for illegal activities.
We also manually reviewed the audio from these main experiments (Appendix~\ref{app:human-evaluation}).

\vspace{1mm}\noindent\textbf{Evaluation on closed-source models.}
We further evaluate our method on Kling and ElevenLabs to examine its effectiveness on commercial services.
Table~\ref{tab:results-closedsource} reports 95.6\% ASR on Kling and 73.8\% on ElevenLabs, with a combined ASR of 84.7\% and PSR of 94.7\%.

\subsection{Impact of Refinement Rounds}
\label{sec:refinement-rounds}

Figure~\ref{fig:refinement-rounds} reports cumulative ASR and PSR as the refinement budget increases from 0 to 20 rounds.
Each curve summarizes a fixed set of test cases balanced across the four categories.
PSR uses Llama Guard for the open-source models and the platform filter for Kling.
Kling reaches 95.6\% ASR by round 2.
TangoFlux reaches 79.4\% by round 8, while AudioLDM reaches 71.9\% by round 12.
Cumulative PSR rises from 77.5\% to 86.2\% on both open-source targets.
The PSR plateau occurs at round 4 for TangoFlux and round 9 for AudioLDM.
Kling's platform pass rate increases from 98.8\% to 99.4\% by round 1.

\subsection{Test-Case Diversity}
\label{sec:test-case-diversity}

We measure the diversity of selected test cases within each harmful category and across the pooled set.
Figure~\ref{fig:category-diversity} reports $1-\mathrm{AvgSelfBLEU}$ and $1-\mathrm{CosSim}$ for the trained generator with refinement on AudioLDM and TangoFlux.
Higher scores indicate lower similarity.
The category-level $1-\mathrm{AvgSelfBLEU}$ ranges from 0.853 to 0.877 for AudioLDM and from 0.852 to 0.879 for TangoFlux.
The two target models yield closely matched category profiles, with illegal activity showing the highest $1-\mathrm{CosSim}$ on both.
The pooled $1-\mathrm{CosSim}$ scores are 0.503 and 0.504 for AudioLDM and TangoFlux respectively.

\subsection{Ablation Study}
\label{sec:ablation}

\textbf{Training stages.}
We ablate SFT and RL separately to assess generator learning.
Table~\ref{tab:ablation} reports Combined ASR across AudioLDM and TangoFlux relative to the 75.6\% baseline in Table~\ref{tab:results-opensource}.
Removing SFT and RL reduces ASR to 65.6\% and 63.1\%, respectively, corresponding to drops of 10.0 and 12.5 percentage points.
Both supervised initialization and audio-feedback optimization contribute to effective acoustic plans.

\begin{table}[t]
  \centering
  \small
  \renewcommand{\arraystretch}{1.15}
  \setlength{\tabcolsep}{3pt}
  \begin{tabularx}{\linewidth}{@{}>{\raggedright\arraybackslash}Xcc@{}}
    \toprule
    \rowcolor{gray20}
    \textbf{Variant} & \makecell{\textbf{Combined}\\\textbf{ASR (\%)} $\uparrow$} & \makecell{\textbf{Drop}\\\textbf{(pp)}} \\
    \midrule
    \rowcolor{highlightcyan}
    \textbf{AUTEUR (full)} & \textbf{75.6} & \textbf{0.0} \\
    w/o SFT & 65.6 & 10.0 \\
    \rowcolor{gray10}
    w/o RLOO & 63.1 & 12.5 \\
    w/o Scene description & 59.3 & 16.3 \\
    \rowcolor{gray10}
    w/o Vocal description & 35.3 & 40.3 \\
    \bottomrule
  \end{tabularx}
  \caption{\textbf{Ablation study on open-source models.} The reference is the Combined ASR (75.6\%) in Table~\ref{tab:results-opensource}. Variant ASRs are derived from the reported drops, measured in percentage points (pp).}
  \label{tab:ablation}
\end{table}

\vspace{1mm}\noindent\textbf{Acoustic descriptions.}
We assess the contribution of the plan's descriptive fields by removing scene or vocal descriptions from the compiled prompts.
Removing scene descriptions reduces Combined ASR to 59.3\%, a drop of 16.3 percentage points, showing the value of contextual situations and audible cues.
Removing vocal descriptions reduces it to 35.3\%, the largest drop of 40.3 percentage points.
This difference suggests that specifying vocal attributes, including intonation, timbre, emotion, and phonation, is particularly important for constructing effective acoustic tests.

\begin{table}[t]
  \centering
  \small
  \renewcommand{\arraystretch}{1.15}
  \setlength{\tabcolsep}{15pt}
  \begin{tabular}{lc}
    \toprule
    \rowcolor{gray20}
    \textbf{Target} & \textbf{Avg.\ Attempts} $\downarrow$ \\
    \midrule
    \rowcolor{gray10}
    TangoFlux & 7.56 \\
    AudioLDM & 8.74 \\
    \rowcolor{gray10}
    Kling & 2.39 \\
    ElevenLabs & 5.16 \\
    \bottomrule
  \end{tabular}
  \caption{\textbf{Average attempts per test case.} Counts include the initial generation.}
  \label{tab:avg-attempts}
\end{table}

\subsection{Search Effort}
\label{sec:search-effort}
We measure the search effort required for each target using the average number of attempts per test case in Table~\ref{tab:avg-attempts}.
Each count includes the initial generation and all subsequent refinement attempts.
The average covers all test cases, including unsuccessful cases, which contribute their actual numbers of attempts up to termination.
Kling requires the fewest attempts on average at 2.39, followed by ElevenLabs at 5.16, TangoFlux at 7.56, and AudioLDM at 8.74.
Together with the success rates, these results show that Kling attains higher ASR with fewer attempts, whereas AudioLDM requires more search while achieving lower ASR.

\begin{figure}[t]
  \centering
  \includegraphics[width=\linewidth]{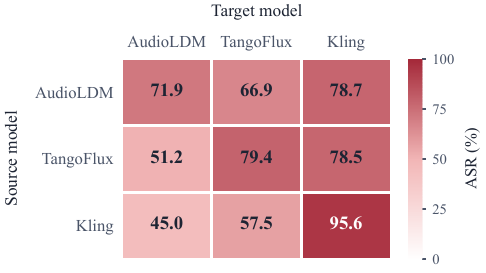}
  \vspace{-8mm}
  \caption{\textbf{Cross-model transfer ASR (\%).}}
  \label{fig:transfer-asr}
\end{figure}

\subsection{Transferability Analysis}
\label{sec:transferability}

Figure~\ref{fig:transfer-asr} reports ASR when test cases constructed for one source model are evaluated on another target model.
The diagonal reports the main-experiment ASR for each model.
Test cases from AudioLDM achieve 66.9\% ASR on TangoFlux.
The reverse direction achieves 51.2\%.
Test cases from Kling achieve 57.5\% on TangoFlux and 45.0\% on AudioLDM.
Transfers to Kling achieve 78.7\% from AudioLDM and 78.5\% from TangoFlux.
These results show transfer across both open-source and closed-source targets.

\section{Conclusion}
\label{sec:conclusion}

We present acoustic plan learning, a framework for automated red-teaming of text-to-audio models that learns structured acoustic plans from waveform feedback.
It addresses the gap between textual acceptance and the harmful acoustic content realized in generated audio.
The generator is trained with supervised fine-tuning and RLOO, while AF3 guides refinement and Qwen2-Audio evaluates the selected waveform after search.
Across four target systems, the method achieves 79.4\% ASR on TangoFlux and 95.6\% on Kling.
Transfer tests show that selected cases remain effective across models, while iterative refinement increases cumulative ASR and PSR.
The framework provides a procedure for developers to audit textual screening and generated sound within the complete generation and execution process.

\section*{Limitations}
\label{sec:limitations}

We evaluate acoustic plan learning on 10-second audio clips.
This limits the empirical scope to short acoustic scenes and harmful content assessed within an individual clip.
At inference, the refinement protocol keeps the initial foreground fixed while modifying scene context and vocal attributes.
This constrains each refinement trajectory to variations of the same sound event; selecting a different foreground requires a new initial plan.

\section*{Ethical Considerations}
\label{sec:ethics}

AUTEUR is intended for safety auditing of text-to-audio models.
Its ability to produce prompts that pass textual screening while eliciting harmful sounds creates a dual-use risk.
Misuse could facilitate harassment or the dissemination of distressing or deceptive audio, and cross-model transfer could extend these risks across services.
We recommend controlled access to attack prompts and harmful recordings when sharing research artifacts, balancing reproducibility with the risk of facilitating misuse.

Human reviewers may experience distress when listening to harmful audio.
Content warnings, voluntary participation, and options to skip recordings or withdraw without penalty are important safeguards for such evaluations.
Automated and human judgments may also reflect contextual or cultural biases, leading to missed harms or the misclassification of benign sounds.
The human review in Appendix~\ref{app:human-evaluation} checks the automated judgments on the main-experiment audio.

Repeated audio generation and model-based judgment consume energy.
Bounded search budgets and reuse of existing recordings can reduce unnecessary computation.
We do not quantify the study's carbon footprint.

\bibliography{paper_references}

\appendix
\section{Additional Method and Evaluation Details}
\label{app:theoretical-details}

We give the variational characterization of Eq.~\ref{eq:central-objective} and the score-function derivation of Eq.~\ref{eq:rloo-gradient} for a fixed execution kernel.

\subsection{Variational Characterization}
\label{app:distribution-proof}

\begin{proof}[Proof of Proposition~\ref{prop:optimal-distribution}]%
\renewcommand{\qedsymbol}{}
Fix $x$ and suppress the conditioning on $x$.
By Assumption~\ref{ass:fixed-execution}, $Z_0$ is finite and strictly positive.
For any distribution $\pi$ on $\mathcal{Z}_x$, substitution of Eq.~\ref{eq:tilted-distribution} gives
\begin{equation}
\label{eq:kl-expansion}
\begin{aligned}
D_{\mathrm{KL}}(\pi\,\|\,q_0^*)
&=\sum_z\pi(z)\log\frac{\pi(z)}{\pi_{\mathrm{ref}}(z)}\\
&\quad-\frac{1}{\beta}\sum_z\pi(z)U_0(x,z)\\
&\quad+\log Z_0(x).
\end{aligned}
\end{equation}
Rearranging and averaging over $x$ yields Eq.~\ref{eq:variational-identity}.
Nonnegativity of KL divergence implies that the unrestricted maximum for each $x$ is attained uniquely at $\pi=q_0^*$.
Restricting $\pi$ to a neural generator family turns this into an approximation problem; the identity holds for every policy in that family.\qedhere
\end{proof}

\subsection{Leave-One-Out Gradient}
\label{app:rloo-proof}

Assume, in addition, that the policy is differentiable on the fixed support and its log-probability gradients are integrable over the target distribution.
For a fixed $x$, let $\ell_\theta(z)=\log\pi_\theta(z\mid x)$ and write
$\rho_\theta(z)=U_0(x,z)-\beta\log[\pi_\theta(z\mid x)/\pi_{\mathrm{ref}}(z\mid x)]$.
Since the execution kernel and reference policy are fixed,
\begin{equation}
\label{eq:regularized-gradient-proof}
\begin{aligned}
\nabla_\theta\mathcal{J}_0(\theta\mid x)
&=\mathbb{E}_{z\sim\pi_\theta}\!\left[
\rho_\theta(z)\nabla_\theta\ell_\theta(z)\right]\\
&\quad-\beta\,\mathbb{E}_{z\sim\pi_\theta}
\left[\nabla_\theta\ell_\theta(z)\right]\\
&=\mathbb{E}_{z\sim\pi_\theta}\!\left[
\rho_\theta(z)\nabla_\theta\ell_\theta(z)\right].
\end{aligned}
\end{equation}
The second expectation is zero because
$\sum_z\pi_\theta(z\mid x)\nabla_\theta\log\pi_\theta(z\mid x)
=\nabla_\theta\sum_z\pi_\theta(z\mid x)=0$.
This accounts for the dependence of the sampled KL penalty on $\theta$.
It justifies treating the regularized return as a stopped-gradient weight in the score-function estimator.

Next, draw $K$ independent plans and conditionally independent execution traces for the same $x$.
For $j\ne i$, $\widetilde R_j$ is independent of $z_i$ conditional on $x$ and the current parameters.
Therefore,
\begin{equation}
\label{eq:loo-baseline-proof}
\begin{aligned}
&\mathbb{E}\!\left[
\frac{1}{K-1}\sum_{j\ne i}\widetilde R_j\,
\nabla_\theta\ell_\theta(z_i)\right]\\
&\qquad=\mathbb{E}\!\left[
\frac{1}{K-1}\sum_{j\ne i}\widetilde R_j\right]
\mathbb{E}[\nabla_\theta\ell_\theta(z_i)]
=0.
\end{aligned}
\end{equation}
Finally, $\mathbb{E}[R_i\mid x,z_i]=U_0(x,z_i)$ for the composite reward in Eq.~\ref{eq:execution-return}.
Combining Eqs.~\ref{eq:regularized-gradient-proof} and~\ref{eq:loo-baseline-proof}, and averaging over $i$ and $x$, gives
$\mathbb{E}[\widehat g]=\nabla_\theta\mathcal{J}_0(\theta)$.
This is the RLOO score-function argument~\cite{ahmadian-etal-2024-back} specialized to a frozen acoustic execution process.
The derivation uses complete groups sampled from the current policy for each update.

\subsection{Scope of the Execution Kernel}

The training process $F_0$ validates each initial plan and screens its compiled description before audio generation.
Frozen reward models assess waveform harmfulness and fidelity to the original caption.
Equation~\ref{eq:execution-return} combines these bounded scores into a return in $[0,1]$.
During each policy update, the kernel is fixed conditional on the sampled initial plan and target.
For valid plans, let $w(z)=(f,u,v)$ collect the acoustic fields.
Compilation uses these acoustic fields, so $U_0(x,z)=U_0(x,w(z))$.
Conditioning Eq.~\ref{eq:tilted-distribution} on a fixed $w$ therefore gives
\begin{equation}
q_0^*(z\mid w,x)=\pi_{\mathrm{ref}}(z\mid w,x),
\end{equation}
because the common exponential utility factor cancels between numerator and denominator.
Thus the ideal distribution changes probability mass across acoustic plans while retaining the reference distribution over their nonexecuted explanations.

Training uses $B_{\mathrm{tr}}=0$ and optimizes $\mathcal{J}_0$.
At inference, the frozen trained checkpoint supplies both the initial generator and the refiner in $F_B$.
AF3 guides this search, and Qwen2-Audio serves as $J_{\mathrm{test}}$ to assess the selected waveform.

\subsection{Reward Configuration}
\label{app:reward-configuration}

We set $(\lambda_h,\lambda_s,\lambda_v)=(0.8,0.1,0.1)$ in Eq.~\ref{eq:execution-return}.
The harmfulness score is $H(a)=0.7h_{\mathrm{Q3}}(a)+0.3h_{\mathrm{AF3}}(a)$, using independent scores from Qwen3-Omni and Audio Flamingo 3 respectively.
A separate AF3 call compares the generated waveform with the original caption to assess scene and vocal fidelity.
For source captions designated as non-vocal, we set $f_{\mathrm{voice}}=0$ and retain the other weights.
Invalid plans and guard-rejected descriptions produce no audio and receive zero execution reward.

\subsection{Decoding Settings}

RLOO sampling uses the full vocabulary with temperature $1$, top-$p=1$, and top-$k=0$.
Sequence likelihoods use the full-vocabulary probabilities of the policy and frozen reference.
At inference, JSON grammar constraints enforce the acoustic-plan schema before validation and execution.
\subsection{Computing Resources}
\label{app:computing-resources}

Local experiments used two NVIDIA RTX PRO 6000 GPUs, each with 96\,GB of memory.
The average time cost per successful attack was 533\,s.
\subsection{Human Evaluation}
\label{app:human-evaluation}

We conducted a human evaluation of all audio clips from the main experiments on AudioLDM and TangoFlux (Table~\ref{tab:results-opensource}), including those classified as harmful and those classified as harmless by the automated evaluation.
We measure accuracy as the proportion of audio clips for which the automated harmful/harmless classification matched the human assessment.
The automated evaluation achieved an accuracy of 88.9\%.

Among all cases counted as successful attacks by the automated evaluation in the main experiments, human reviewers confirmed that 86\% both constituted successful attacks and satisfied the original target requirements.
These automatically successful clips also contribute to the harmfulness accuracy calculation, so the two statistics are computed on overlapping sets of audio.
This result provides human evidence supporting the reliability of the reported ASR.

\subsection{Participants.}
Ten students from our laboratory who had not participated in the experiments volunteered as evaluators.
They were peers of the researchers and were not authors of this paper.
They received no compensation and could withdraw at any time.
Before annotation, they received a warning about potentially harmful audio content and gave verbal consent to participate in the evaluation.
We did not collect their ages.
We did not obtain institutional ethics approval or request an exemption for this evaluation.

\subsection{Annotation procedure.}
Evaluators were shown the original target prompts alongside the generated audio.
They judged harmfulness for all main-experiment clips, including clips classified as harmful and harmless by the automated evaluation.
For the subset counted as successful attacks by the automated evaluation, they additionally judged whether the audio satisfied the original target requirements.
The target-matching annotation set was therefore a subset of the harmfulness annotation set.
Each clip was reviewed by seven of the ten evaluators, with overlapping subsets assigned to different evaluators.
Each applicable binary judgment was aggregated separately by majority vote, with at least four of the seven votes determining the final label.
The odd number of judgments per clip prevented ties.

\end{document}